# Fusion of multispectral satellite imagery using a cluster of graphics processing unit

Al-Oraiqat A.[*], Bashkov E.[**], Babkov V.[**], Titarenko C.[**]

[*]Taibah University, Community College, Department of Computer Sciences & Information
Kingdom of Saudi Arabia, p.o.Box 2898
anas_oraiqat@hotmail.com

[**]Donetsk National Technical University, 83000, Donetsk, Artyoma 58, Ukraine
victor.babkov@gmail.com

**Abstract**

The paper presents a parallel implementation of existing image fusion methods on a graphical cluster. Parallel implementations of methods based on discrete wavelet transformation (Haar's and Daubechie's discrete wavelet transform) are developed. Experiments were performed on a cluster using GPU and CPU and performance gains were estimated for the use of the developed parallel implementations to process satellite images from satellite Landsat 7. The implementation on a graphic cluster provides performance improvement from 2 to 18 times. The quality of the considered methods was evaluated by ERGAS and QNR metrics. The results show performance gains and retaining of quality with the cluster of GPU compared to the results obtained by the authors and other researchers for a CPU and single GPU.

1. **Introduction**

Remote sensing is a process of gathering information about the object, area or phenomenon without direct contact. Methods of remote sensing founded on analog or digital registration based on electromagnetic emissions from objects or on reflected waves over a wide spectral range. The products of remote sensing are digital images with pixel values corresponding to the value(s) of properties (such as reflectivity, temperature, etc.) in respective areas. The size of the area depends on the resolution of the images.

Most satellites such as SPOT, IRS, Landsat 7, IKONOS, Quick Bird, Orb View, and sensors, such as the Leica ADS40, provide panchromatic images with high resolution and multispectral with low resolution [1]. An effective method for merging images may expand the use of such images in many areas that require high spectral and spatial image resolution simultaneously.

Thus, the task of merging remote sensing results is the integration of geometric details from panchromatic (PAN) image with a high resolution and color information contained in multispectral (MS) image with a lower resolution (for example in geo-information systems) [2].

Since 80's of the 20[th] century was developed a number of methods for image fusion for remote sensing problems [3, 4]. This work is related to image fusion for remote sensing and has the main goal of improving productivity of image fusion process. Recently, Graphics Processing Units (GPUs) are widely used for high performance computing, including image fusion [5, 6].

Graphical cluster is a distributed parallel computing system with nodes supporting general-purpose computing, equipped with video cards. This paper describes the implementation methods of image fusion using a cluster of GPU.

2. **The choice of image fusion method**

Techniques of image fusion can be divided into such classes: additive and multiplicative methods and methods based on filters or the relationships between the channels and transformations such as the discrete wavelet transformation (DWT) [7, 8].

DWT-method based on replacing the coefficients of the approximations with the data from multispectral channel was chosen as a main method for the parallel implementation. The choice of this method is grounded by high expectations of metrics value of the fusion quality. Fig. 1 shows the explanation of this method.

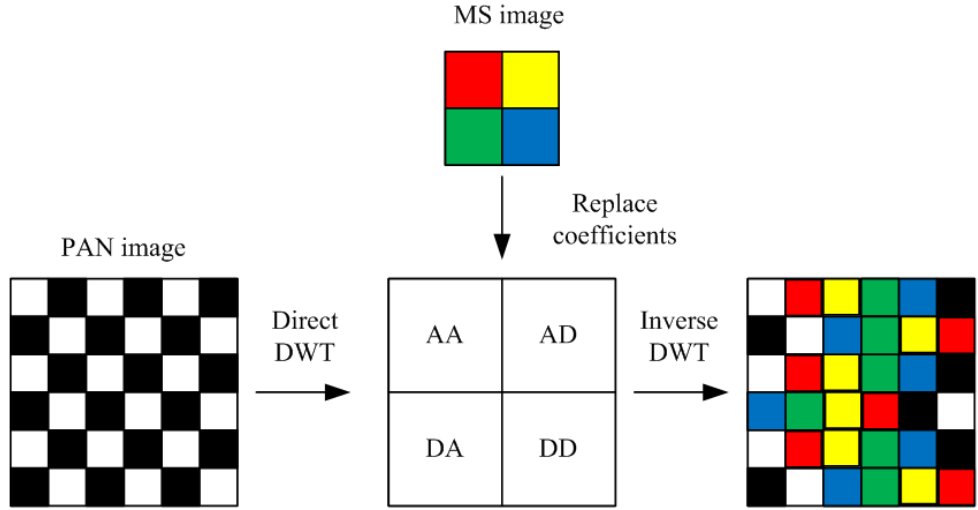

Figure 1: DWT-based methods explanation.

Ratio of the channels size (PAN and MS) is assumed to be 4/1. The method contains these steps:
- Direct 2D DWT of the PAN channel.
- Replacement of coefficients of the approximation by the multispectral data channel.
- Inverse 2D DWT.

The 2D transform is performed by applying a one-dimensional transform to the rows of the matrix representing the image and then to the columns. Similarly, two-dimensional inverse transform is performed by applying a reverse one-dimensional transform to the columns and then to the rows of the matrix.

The paper shows the implementation of two methods based on DWT–Haar and Daubechie methods [9, 10]. The direct 1D discrete wavelet Haar transform is performed by Eq.(1)-(2):

$$s'_i = (s_{2i} + s_{2i+1}) / 2 \tag{1}$$

$$s'_{N/2+i} = (s_{2i} - s_{2i+1}) / 2 \tag{2}$$

where $i \in [0, N/2]$, $s_i$ is the component of the input vector, $s'_i$ is the component of the output vector, and $N$ is the vector length.

The direct 1D discrete wavelet Daubechies transform is performed by Eq. (3)-(6):

$$s'_i = \sum s_{2i+k} h_k \tag{3}$$

$$s'_{N/2+i} = \sum s_{2i+k} g_k \tag{4}$$

$$s'_{N/2-1} = \sum s_{N-2+m} h_m + \sum s_m h_{2+m} \tag{5}$$

$$s'_{N-1} = \sum s_{N-2+m} g_m + \sum s_m g_{2+m} \tag{6}$$

where $i \in [0, N/2 - 1]$, $k \in [0, 3]$, $m \in [0, 1]$, $s_i$ is the component of the input vector, $s'_i$ is the component of the output vector, $N$ is the vector length, and $h$ and $g$ are calculated by Eq. (7)-(10):

$$h_i = (1 + 2i + \sqrt{3}) / 4\sqrt{2} \tag{7}$$

$$h_{i+2} = (3 - 2i - \sqrt{3}) / 4\sqrt{2} \tag{8}$$

$$g_i = (2i + \sqrt{3} - 3) / 4\sqrt{2} \tag{9}$$

$$g_{i+2} = (\sqrt{3} - 1 - 2i) / 4\sqrt{2} \tag{10}$$

where $i \in [0, 1]$. The inverse 1D discrete wavelet Haar transform is performed by Eq. (11)-(12):

$$s_{2i} = s'_i + s'_{N/2+i} \quad (11)$$

$$s_{2i+1} = s'_i - s'_{N/2+i} \quad (12)$$

where $i \in [0, N/2)$. The inverse 1D discrete wavelet Daubechies transform is performed by Eq. (13)-(16):

$$s_0 = \sum s'_{N/2-1+k} t_{3k} + \sum s'_{(N-1)k} t_{2-k} \quad (13)$$

$$s_1 = \sum s'_{N/2-1+k} u_{3k} + \sum s'_{(N-1)k} u_{2-k} \quad (14)$$

$$s_{2+2i} = \sum s'_{i+k} t_{2k} + \sum s'_{i+N/2+k} t_{1+2k} \quad (15)$$

$$s_{3+2i} = \sum s'_{i+k} u_{2k} + \sum s'_{i+N/2+k} u_{1+2k} \quad (16)$$

where $i \in [0, N/2 - 1]$, $k \in [0, 1]$, and t and u are calculated as $t_0 = h_2$, $t_1 = g_2$, $t_2 = h_0$, $t_3 = g_0$, $u_0 = h_3$, $u_1 = g_3$, $u_2 = h_1$, $u_3 = g_1$. $u_i, g_i$ are obtained using Eq. (7)-(10).

### 3. Methods implementation on parallel system

Assuming that the result of the fusion of parts followed by a union is equivalent to fusion of the original data set as a whole, can offer a scheme for task division on the cluster nodes. The source matrix is split horizontally and vertically into equal parts, each part is then processed on a separate node in the cluster and the final results merge into a single image on the master node. Each cluster node performs image processing, pixel by pixel, using two-dimensional array of threads. A schematic of the partitioning process is shown in Fig. 2. Fig. 3 shows a flowchart of the parallel algorithm for multispectral fusion.

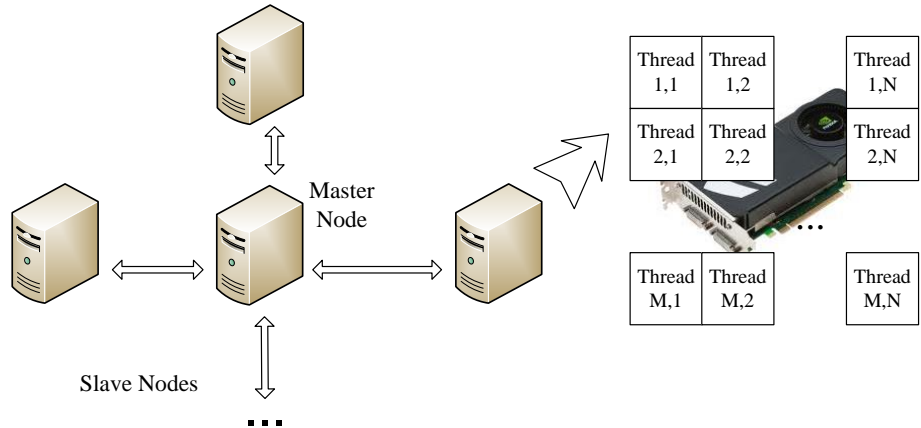

Figure 2: Schematic of graphical cluster's.

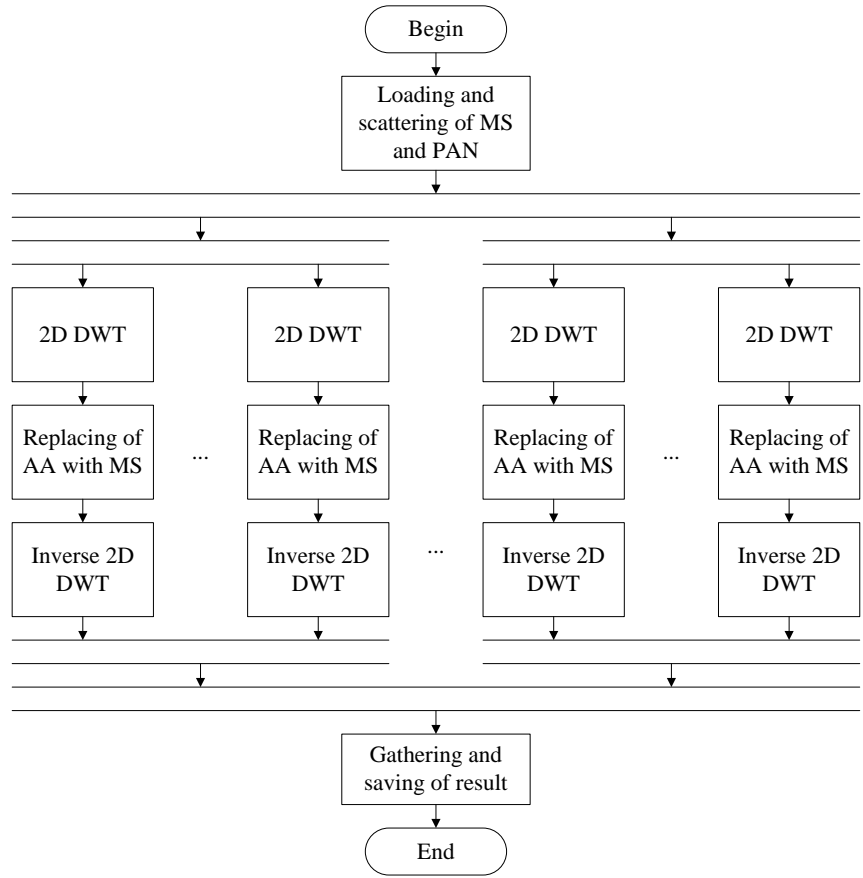

Figure 3: Flowchart of parallel algorithm for multispectral image fusion.

To organize the communication between nodes in the cluster, MPI implementation from Argonne National Laboratory - MPICH2 was used [11]. Image processing on each node is performed using General-Purpose Computing on Graphics Processing Unit (GPGPU) card with a Compute Unified Device Architecture (CUDA) technology [5].
In order to accelerate the process of transferring data between nodes, images are transferred in to the representation of 8 bpp. However, at a fusion, the processes are performed using the floating point single precision representation of 32 bpp which is necessary for the discrete wavelet transform.

The fusion process on the GPGPU card consists of the following steps for each channel:
- load PAN channel in the first block in texture memory;
- allocate an array in global memory for the PAN channel;
- load MS channel in the second block in texture memory;
- perform direct 1D DWT over the rows of PAN array with reading data from texture first block and writing into the global array;
- copy pixels from global array to texture first block;
- perform direct 1D DWT over the columns PAN array with reading data from texture first block and writing into global array;
- replace the coefficients of approximation in global array with MS coefficients from texture second block;
- copy global array to texture first block;
- perform inverse 1D DWT over columns reading from texture first block and writing into the global array;
- copy global array to texture first block;
- perform inverse 1D DWT over rows reading from texture first block and writing into the global array;
- unload the result from global array.

In this case, it is expected to achieve a high performance through the reading from textures' memory and sequential writing into the global memory and copying data between components of devices without using a temporary memory.

**4. Experimental results**

Experiments were performed on a graphic cluster which consists of four nodes with the following characteristics:
- Intel Core 2Quad Q8400;
- NVIDIA GeForce GTX460 video card;
- RAM (DDR3) 4 GB;
- PCI network card(1Gbit/s);
- Operating system: Windows 7 Professional x64.

To provide the basis for comparing the quality and performance of methods, two other methods for implementation have been chosen: a weighted averaging method and a method based on the transformation of color space (IHS) [7, 8]. Also, for all implemented methods, metrics have been obtained for the cluster of CPUs.

For the experiments on real data service USGS Global Visualization Viewer from U. S. Geological Survey were used. To obtain estimates of productivity and quality, imagery were selected, from satellite Landsat 7 [12], which are area pictures of the Donetsk region (Ukraine).

In table 1and 2 are used followed abbreviators:
- WA – weighted averaging method;
- IHS – Intensity Hue Saturation method;
- HDWT – Haar discrete wavelet transform;
- DDWT – Daubechies discrete wavelet transform.

As follows from Table 1, the implementation of the IHS method has the best time on cluster with GPU. This is because the GPU instruction set is more efficient for implementation of IHS.

As follows from Table 2, the implementation of the HDWT method has the best time on cluster with CPU. This is because this implementation, efficiently implemented using arithmetic operations of CPU, does not use image scaling and discrete wavelet transform.

Table 1: Experimental Results for Cluster of GPU.

| PAN width, px | PAN height, px | $T_{WA}$, sec | $T_{IHS}$, sec | $T_{HDWT}$, sec | $T_{DDWT}$, sec |
|---|---|---|---|---|---|
| 16280 | 14960 | 3,73 | 3,01 | 3,61 | 3,70 |
| 8140 | 7480 | 0,90 | 0,85 | 0,98 | 1,02 |
| 4070 | 3736 | 0,32 | 0,30 | 0,34 | 0,35 |

Table 2: Experimental Results for Cluster with CPU.

| PAN width, px | PAN height, px | $T_{WA}$, sec | $T_{IHS}$, sec | $T_{HDWT}$, sec | $T_{DDWT}$, sec |
|---|---|---|---|---|---|
| 16280 | 14960 | 49,0 | 67,1 | 11,6 | 13,5 |
| 8140 | 7480 | 2,08 | 6,24 | 2,89 | 3,23 |
| 4070 | 3736 | 0,51 | 1,57 | 0,66 | 0,73 |

Because the goal of this work was to improve the performance of parallel implementations, we should determine how to measure the performance. In this case, the output will be the number of operations per second, whereas operations with memory and assuming that all operations are on average equal in complexity.

Fig.4 shows the comparison of growth performance for each method among implementations on graphic cluster and cluster of CPU. Based on analysis of Fig.4, we can conclude that the performance gain of parallel implementations on GPU lies in the range 2-18 times depending on the fusion method. In order to illustrate and clearly differentiate among the considered methods, Fig.5 shows four images that are the results of each of the implemented methods.

The metrics ERGAS (Error Global in Synthesis) and QNR (Quality Non Reference) are used for evaluating the quality of the considered methods (Table 3).

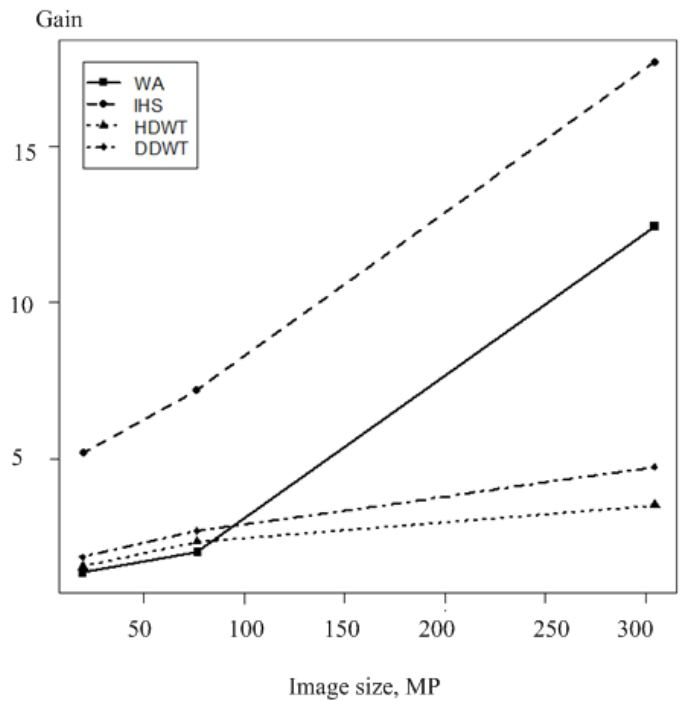

Figure 4: Performance improvement of cluster with GPU compared to the cluster of CPU.

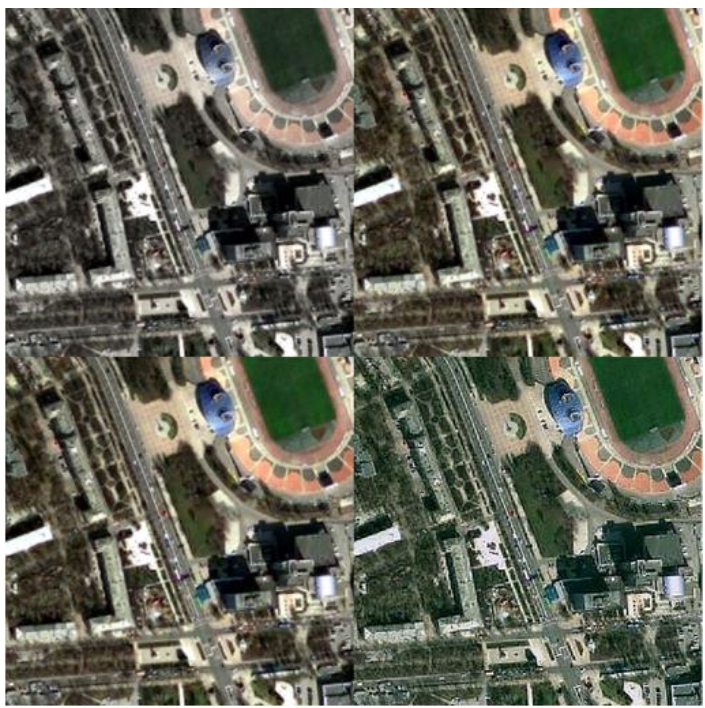

Figure 5: Result of image fusion (from left to right and from top to bottom - WA, IHS, HDWT, and DDWT).

Table 3: Comparison of quality metrics.

| Metric | WA | HIS | HDWT | DDWT |
|--------|------|------|------|------|
| QNR | 0.35 | 0.30 | 0.42 | -0.10 |
| EGRAS | 50.85 | 42.98 | 27.43 | 79.84 |

As follows from Table 3, the HDWT method has the best quality (smallest error for ERGAS, and the highest quality for QNR), that consistently agrees with the subjective visual assessment of the result images of Fig.5. It should be noted that metrics for DDWT method do not meet the expected results. Based on theoretical assumptions, the DDWT method should provide a highly detailed image which is associated with high quality metrics (QNR). But the results of the experiment show that the DDWT has a negative QNR. This may be explained by "over injection" and as a result we have negative QNR (spatial quality index is higher than the spectral so their difference is negative).

A performance gain in the range of 2-18 (Figure 4, Table 1 and Table 2) could be compared to similar results that were obtained in [13]. In that reference, the authors obtained 10% increase in the performance using GPU compared to using CPU. They used single GPU GTX 550Ti. This research shows that the GPU cluster even with a less powerful processor GTX 460 could work in 2-18 times faster with the same image size.

5. Conclusion

The paper presents a parallel implementation of existing methods of fusion of images on the architecture of a graphic cluster. Experimental results of testing of parallel implementations show that the IHS method has the best result on a cluster of CPU and the HDWT method has the best result on a cluster of GPU. The implementation on a graphic cluster provides performance improvement from 2-18 times. The proposed parallel implementation used for image fusion in a software system for processing remote sensing data from satellite [13].

**Acknowledgement**

The authors would like to aknowledeg the coperation of the Computer Sience and Information Deparments of Taibah Univesrity, Madina, KSA, and Donetsk National Technical University, Ukraine for providing technical facilities that have been used during the development of this research. The authors would also greatly appreciate the valuable comments of the anonymous reviewers that helped in improving the presentation of the final version of the paper.

**References**

[1] Zhang Y., Understanding Image Fusion, Photogrammetric Engineering & Remote Sensing, 70(6), 657 (2004).
[2] Simone G., Farina A., Morabito F.C., Serpico S.B., Bruzzone L., Image fusion techniques for remote sensing applications, Information Fusion Journal, 3, 3 (2002).
[3] Bretschneider T., Kao O., Image Fusion in Remote Sensing, Online Symposium for Electronics Engineers, ISSN 1532-513X , (2000).
[4] Aiazzi B., Baronti S., Selva M., Improving component substitution Pan-sharpening through multivariate regression of MS + Pan Data, IEEE Trans. Geosci. Remote Sensing, 45(10), 3230 (2007).
[5] Sanders J., Kandrot E., CUDA by Example: An Introduction to General-Purpose GPU Programming, Addison-Wesley, ISBN-10: 0131387685, (2011).
[6] Yoo S., Jo S., Choi K., Jeong C., A Framework for Multisensor Image Fusion using Graphics Hardware, 11- the International Conference on Information Fusion, 1 (2008).
[7] Pajares G., de la Cruz J.M., Wavelet-based image fusion tutorial, Pattern Recognition, 37(9), 1855 (2004).
[8] Rahmani S., Strait M., Merkurjev D., Moeller M., Wittman T., An Adaptive IHS Pan-sharpening Method, IEEE Letters on Geoscience and Remote Sensing, 7(4), 746 (2010).
[9] Beylkin G., Coifman R., Rokhlin V., Wavelets in Numerical Analysis, In Wavelets and Their Applications, New York, Jones and Bartlett, ISBN-10: 0867202254, 181 (1992).
[10] Kingsbury N., Magarey J., Wavelet Transform in Image Processing, Proc. First European Conference on Signal Analysis and Prediction, 24 (1997).
[11] http://www.mcs.anl.gov/research/projects/mpich2/.
[12] http://glovis.usgs.gov/.
[13] Jeong I., Hong M., Hahn K., Choi J., Kim C., Performance Study of Satellite Image Processing on Graphics Processors Unit Using CUDA, Korean journal of remote sensing, 28(6), 683 (2012).